\documentclass[10pt,twocolumn,letterpaper]{article}

\usepackage{wacv}
\usepackage{times}
\usepackage{epsfig}
\usepackage{graphicx}
\usepackage{amsmath}
\usepackage{amssymb}
\usepackage{multirow}

\usepackage{algorithm}
\usepackage[noend]{algpseudocode}
\usepackage[caption=false]{subfig}
\usepackage{mathtools}
\usepackage{tabularx}
\usepackage[accsupp]{axessibility}

\DeclareMathOperator*{\argmax}{arg\,max}

\def\wacvPaperID{762} 

\wacvfinalcopy 

\ifwacvfinal
\fi

\ifwacvfinal
\usepackage[breaklinks=true,bookmarks=false]{hyperref}
\else
\usepackage[pagebackref=true,breaklinks=true,colorlinks,bookmarks=false]{hyperref}
\fi

\begin{document}

\title{RLSS: A Deep Reinforcement Learning Algorithm for\\Sequential Scene Generation}

\author{Azimkhon Ostonov, Peter Wonka, Dominik L. Michels\\
KAUST Visual Computing Center\\
Campus, Bldg 1, Thuwal 23955, KSA\\
{\tt\small \{azimkhon.ostonov, peter.wonka, dominik.michels\}@kaust.edu.sa}
}

\maketitle
\thispagestyle{empty}

\begin{abstract} We present RLSS: a reinforcement learning algorithm for sequential scene generation. This is based on employing the proximal policy optimization (PPO) algorithm for generative problems. In particular, we consider how to effectively reduce the action space by including a greedy search algorithm in the learning process. Our experiments demonstrate that our method converges for a relatively large number of actions and learns to generate scenes with predefined design objectives. This approach is placing objects iteratively in the virtual scene. In each step, the network chooses which objects to place and selects positions which result in maximal reward. A high reward is assigned if the last action resulted in desired properties whereas the violation of constraints is penalized. We demonstrate the capability of our method to generate plausible and diverse scenes efficiently by solving indoor planning problems and generating Angry Birds levels. \end{abstract}

\section{Introduction}

Generative modeling has seen drastic improvements in recent years. Especially for images, state-of-the-art generative adversarial networks (GANs) produce fantastic results~\cite{goodfellow2014generative,stylegan,styleganv2}. Even though generative models such as GANs or variational autoencoders (VAEs) \cite{kingma2013autoencoding} are considered to be unsupervised methods, a large amount of data is still required.
While multiple attempts have been made in reproducing the success of image-based generative models in other domains, such as scenes, meshes, and point clouds, the results are far behind.
A major obstacle is the lack of data, as there are no high quality data sets of scenes that contain many models. Many data sets are small, e.g.~\cite{article_talton}, or contain too many low quality models. For example, the SUNCG data set~\cite{inproceedings2} contains multiple low quality models generated by amateur modelers that violate commonly accepted hard constraints (in addition, the data set is currently unavailable due to a legal dispute).

In order to make progress on generative modeling for scene generation, we propose to build a learning framework that does not require a large amount of training data. In this context, reinforcement learning is an ideal choice as the need for training data is replaced by reward function design. As reinforcement learning is traditionally applied to maximize expected cumulative reward, it is not generally used to generate a large variety of scenes. In this work we propose a reinforcement learning algorithm to be able to generate a large variety of scenes. We call our approach RLSS. It operates sequentially and places scene objects one-by-one.

Specifically, our major contributions are as follows.
\begin{itemize}
    \item We present RLSS: (to our knowledge) the first reinforcement learning based scene generation algorithm. The distinguishing characteristic is that RLSS can generate a large variety of scenes for the same input (i.e. scene boundary) taking into account domain constraints.
    \item We propose an efficient approach for scene synthesis problem which enables to solve this task by separating the problem requirements into two different categories: hard constraints that are included in the environment and predefined design objectives, which the network learns how to achieve, given different initial scene boundaries, during the training process.
    \item Our reward designing approach makes it easy to adapt this method for many scene synthesis problems without spending too much time on reward calculation for actions.
    \item We demonstrate the advantages of our method to generate plausible and diverse scenes efficiently by solving indoor planning problems and generating Angry Birds levels. 
\end{itemize}

\section{Related Work}

Employing reinforcement learning in the context of generative problems has been addressed in the community.\\
\indent SPIRAL \cite{ganin2018synthesizing} is a reinforcement learning based adversarial agent which learns to synthesize visual programs for graphic engines in order to generate images. It has been shown that their method works well for image reconstructions on MNIST \cite{lecun2010mnist} and OMNIGLOT \cite{article_lake} data sets but fails to synthesize new examples. Model-based reinforcement learning \cite{Huang_2019} was applied to image reconstruction for stroke-based paintings. The fundamental difference between these methods and our method is that in contrast to rewarding the agents based on the discriminators' outputs, in our case this is done by simulating the environment and checking relevant constraints.\\
\indent Formalizing reinforcement learning as a probabilistic inference technique has been explored in  \cite{dayan1997,abdolmaleki2018,levine2018reinforcement}. Connecting reinforcement learning with posterior regularization \cite{article_pr} to incorporate the domain constraints to deep generative models has been considered in \cite{hu2018deep}. Compared to this approach, our method is based on directly employing reinforcement learning as a generative model.\\
\indent Deep generative models \cite{goodfellow2014generative, kingma2013autoencoding} have been extensively used by scientific community to synthesize scenes in indoor planning \cite{article2, article_grains, article_planit, Ritchie_2019, articleZhang}. These models are mostly image-based and therefore use pixel-level reasoning to distinguish objects and extract spatial features from an image. Which can be poorly adapted when handling various domain-based constraints in scene synthesis problems. As a result, one either need to generalize the problem and accept more simplified domain constraints \cite{article_grains}, or  use constraint violation check at each step along with generative models to deal with the problem \cite{article2, article_planit}. However, this is associated with costs: in the first case, we will solve the sub-problem instead of solving the real problem, in the second case, this leads to an increase in the synthesis time.\\
\indent Another major direction to generate indoor scenes is related with using Markov chain Monte Carlo (MCMC) methods \cite{inproceedings_human_centric, articleMerrell, articleMakeitHome, Kermani2016}. In these methods problem specific objective functions are optimized based on some rule based criteria. The main drawback of using MCMC sampling in these methods is generation time, which requires thousands of iterations to synthesize one plausible scene.\\
We also would like to mention the reinforcement learning has been used in several contexts different from generative problems~\cite{DBLP:journals/corr/abs-1803-01129,DBLP:journals/corr/abs-1708-05884, DBLP:journals/corr/ZophL16, DBLP:journals/corr/abs-1912-03966, HU2021101878}.

\section{Method}
In this section, we provide an overview of RLSS: our reinforcement learning based sequential scene generation algorithm. We provide required background information and discuss relevant components in detail.

\subsection{Overview}\label{section:3.1.}
We developed RLSS based on the proximal policy optimization (PPO) \cite{schulman2017proximal} method for scene generation problems. The scene we would like to generate is represented as an environment in which objects are added iteratively one-by-one. The neural network is trained on a 2D representation of the scene. If the visual content we are generating has three dimensions, then its most informative 2D view is used during the training. The set of all actions which can be applied in the environment is denoted with $A$. The reward which the agent obtains after taking an action $a \in A$ is denoted by $r$. This formulation of the scene allows us to use reinforcement learning algorithms. However, the use of standard reinforcement learning is limited to optimization problems, where the agent needs to collect as much reward as possible in a virtual environment. In each step the agent chooses an action among the set of possible actions which at the end will result in the highest reward. Therefore, in the standard reinforcement learning setup, the main target of the agent is to maximize the cumulative reward. In our setting we additionally consider increasing variety, which is achieved by introducing randomness in the action selection process.\\
\indent We introduce the following two problem-specific key definitions which will help us to effectively explain our method.
\begin{itemize}
    \item \textbf{Hard constraints} are a set of constraints that should always be satisfied when synthesizing a scene. Usually, this type of constraints come from the problem domain and violating these constraints makes it impossible to continue the generation process. In our method, these constraints are included in the environment.
    \item \textbf{Design objectives} comprise a set of requirements which shape the general appearance of the generated scene. Each successfully generated scene should have these features. This can be understood as a broad definition of a scene we would like to synthesize. Only such scenes which fulfill predefined design objectives could be considered as valid examples for good scenes.
\end{itemize}
\indent \textit{Different from the most of the scene synthesis algorithms our method learns how to generate successful scene based on predefined criteria (design objectives) and domain knowledge is applied to place objects in the scene.} This method has particular advantages where relationship or rules exist for placing objects. For example, in physics based games or in indoor planning.  

\subsection{Scene Abstraction} \label{section:3.4a}
A scene consists of the initial scene boundary and set of objects, each defined with its center's position $P(x, y, z)$, bounding box (box with minimal volume which can fully contain this object) and orientation $\alpha$ relative to local coordinate axis.\\
\indent \textit{Structures (groups).} First, we find independent structures in the problem domain. 
Each structure consists of one or more objects and together these objects play some function in the current problem domain. These structures can be derived from acceptable scene examples. We define the set of all structures we use in the problem as $\mathcal{S}$. Each structure $st\in\mathcal{S}$ has a complexity (the number of different objects in this structure) defined as $C(st)$.\\
\indent \textit{Placements}. Many structures have similar arrangements in the scene. For example, table and chair, dresser and ottoman, sofa and coffee table can be placed in a similar way. For a given set of objects we define set of possible placements $\mathcal{P}$ to place these objects in the scene. Objects in the placements are defined by their top left coordinate of the bounding box and orientation. These placements are extracted from acceptable scene examples. Objects are placed in the scene according to one of the placements which makes generated scenes more realistic.

\subsection{Basic Network and State Representation}\label{section:3.2.}
Reinforcement learning (RL) learns how to control an agent in an environment, to maximize cumulative reward. Given an observation $o_t$ at time $t$, the agent performs an action $a_t$ according to its policy $\pi$ and receives a reward $r_t$ for the current action. The policy is a mapping between state $s_t$, the environment's current condition to the action. The environment can be described as a Markov Decision Process, i.e., the current state of the environment fully characterizes the process. The total reward from time step $t$ is defined as $$R_t = \sum_{i=0}^{\infty} \gamma^i r_{t+i}(s_i, a_i)\,,$$ $\gamma \in (0, 1]$ in this regard is a discount factor which indicates how important are the future rewards at the current point.

\indent Policy based methods are based on adjusting policy in order to maximize the expected reward. We use PPO \cite{schulman2017proximal} with actor-critic network with shared parameters between the policy $\pi (a_t|s_t;\theta)$ and value function $V(s_t; \theta_v)$. The policy is updated with clipped policy gradient objective $$L^{CLIP} (\theta) = \mathbb{\hat{E}}_t\left[\min(r_t(\theta)\hat{A_t}, clip(r_t(\theta), 1-\epsilon, 1+\epsilon)\hat{A_t})\right],$$ where $\hat{A_t}$ advantage function estimator and $r_t(\theta) = ({\pi (a_t|s_t;\theta)})/({\pi (a_t|s_t;\theta_{old})})$ probability ratio.
The final objective term includes value updates and entropy bonus to encourage exploration \cite{schulman2017proximal} $$L_t(\theta) = \mathbb{\hat{E}}_t\left[L^{CLIP}_t (\theta)-c_1 L^{VF}_t (\theta)+c_2 S[\pi (a_t|s_t;\theta)](s_t)\right].$$
The value function ($L^{VF}_t (\theta)$) is updated with squared-error loss between the value function output and the total reward from time step $t$.

\indent\textbf{State} \ consists of: 2D representation of the scene which includes scene boundary and objects; object existence per category indicates whether the scene has at least one object from this category, takes value from $\{0, 1\}$; object availability per category which is equal to $1$ if number of objects is less than maximal allowed $0$ otherwise; scene condition; step indicator.

\subsection{Action Space Separation} \label{section:3.3.}
One main concern which needs to be taken into account when using RL is dealing with a large action space. For the scene synthesis problems action space $A$ formed from the Cartesian product of two sets $O$ and $P$, $A = O \times P = \{(o,p)|o\in O$ and $p\in P\}$. Where $O$ denotes the set of all objects and $P$ denotes the set of all positions to place these objects in the scene. As a result of this, action space could grow rapidly even for a smaller number of objects and positions. Thus creating problems for RL algorithms to learn efficient policies. In our approach we limit action space with objects only, $A=O$. The positions for placing these objects are determined by a greedy algorithm, based on the current state and the object being added to the scene. More concretely, $p_{i} = \argmax_{j \in P} r(s, a, j)$, $p_{i}$ - position for the current object, $r(s, a, j)$ - reward function.

\subsection{Reward Designing} \label{section:3.4.}
\indent At each step the scene generated with our algorithm is checked with respect to the following conditions: successful scene conditions ($CheckSuccessfulCondition()$), failure conditions ($CheckFailureCondition()$), constraints on the number of objects of the same type ($CheckObjectCountCondition()$). Each episode ends with a successful or failed scene generation. If after some steps the generated scene has met the failure condition, then the corresponding reward is $-1$. In contrast, once all of the predefined design objectives are obtained, which means a successful scene generated, then we assign the maximum reward. For both of these cases the scene generation ends and it will start again from the beginning. Otherwise, the reward is calculated depending on the type of design objectives which are achieved taking the last action.\\
\indent In each step, only one type of object $tp \in O$ chosen by the RL algorithm can be placed within the scene. The objects which were already placed in the scene can be formally written as a union of substructures located in the different positions of this scene. A substructure is a part of some structure $st\in \mathcal{S}$. We first find the maximum complexity of substructures existing in the scene ($Search()$), which the current object can be grouped with, while not violating hard constraints. If there are more than one substructures with maximum complexity, then the choice is arbitrarily done between them. Then the current object is grouped with this substructure. The resulting substructure is part of some structure $st^{'}\in \mathcal{S}$. The reward for the placing of this object depends on the complexity of the substructure we can get from this placement. The higher the resulting complexity, the higher is the reward. This encourages the RL algorithm to find optimal policies to get a higher cumulative reward. From the other side, we use only $l$ different positive rewards not depending on the type of the structure, one for each complexity: $r_{1}\leq r_{2} \leq \cdots \leq r_{l}$, which serves to increase the variety of generated scenes. Where, $l = \max\limits_{\displaystyle{st\in \mathcal{S}}} C(st)$ is maximum complexity among all structures. Constraints on the number of elements of the same type are applied by assigning negative rewards. Successful and failed generation conditions depend on the problem domain. All the rewards in our method are taken from the $[-1, 1]$ interval. The generation process is summarized in Algorithm~\ref{Algorithm1}. $PlaceObject()$ function in this scope uses one of the placement functions depending on the state, current object and complexity.\\
\indent \textit{Predefined design objectives.} Structures in the scenes exhibit different relationships between objects. 
As we quantify these relationships with numbers (rewards), we can numerically assess the current condition of the generated scene. In this paper, the sum of all rewards taken so far $\sum_{i=1}^t r_i$ represents the current condition of the scene. From positive examples we can get minimal value $R_m$ for a scene to be considered successfully generated. Also, we can enforce other requirements along with minimal value, depending on the problem. These conditions follow form predefined design objectives.

\algrenewcommand\textproc{}

\begin{algorithm}
\caption{Reward Assigment}
\hspace*{\algorithmicindent} \textbf{Input:} current state, current action, reward vector sorted in descending order \\
\hspace*{\algorithmicindent} \textbf{Output:} reward for current action 
\begin{algorithmic}[1]

\Procedure{AssignReward}{$state, action, r[\;]$}

\If{$CheckFailureCondition(state)$}
\State \Return $-1$
\EndIf
\If{$CheckObjectCountCondition(state)$}
\State \Return $-0.1$
\EndIf

\State {$l$ $\gets$ {$length(r)$}}
\For{$i \gets 1$ to $l$} 
        \State {$c$ $\gets$ {$l-i+1$}} \Comment{Complexity}
        \State {$s$ $\gets$ $Search(state, action, c)$}
        \If{$s.not\_empty()$}
        \State {$p$ $\gets$ {$random(1, length(s))$}}
        \State {$PlaceObject(state, action, p)$}
        \If{$CheckSuccessfulCondition(state)$}
        \State \Return $1+r[i]$
        \Else \State \Return $r[i]$
        \EndIf
\EndIf
\EndFor
\State \Return $-0.1$
\EndProcedure

\end{algorithmic}
\label{Algorithm1}
\end{algorithm}

\subsection{Network Training} \label{section:3.5.}
The main objective of our method is to generate a wide variety of scenes which satisfy hard constraints and at the same time have predefined properties. In each step we add an object to the scene, until we found that the final condition is reached, successful or failed scene generated. We denote this process as an episode, and the scene as an environment.\\
\indent \textit{Action sampling.} 
The objective in standard RL is to maximize $\mathbb{E}[R_t]$ expected cumulative reward. In our setting we additionally would like to increase variety of generated scenes.\\ We normalize $\pi(a_i |s; \theta)$ policy output (unnormalized) for a scene $s$, using the softmax function with temperature $\tau$:\\
\begin{align*}
P^{\tau}(a_{i}|s;\theta) = \frac{e^{\pi(a_i |s; \theta)/\tau}}{\sum_{z=1}^{N}e^{\pi(a_z |s; \theta)/\tau}}\,.
\end{align*}
Here, $P^{\tau}(a_{i}|s; \theta)$ specifies the probability of taking action $a_{i}$ given state $s$. In order to balance between exploration and exploitation, $\tau$ is steadily decreased from $1$ completely randomly to $0$ greedy action sampling. And then we keep greedy action sampling until convergence during the training process.\\
\indent During the inference time we sample actions with some $\tau$ value. 
As $\tau$ close to 1 the results show more variety but the percent of scenes which have predefined properties will be small. Conversely, for the values of $\tau$ close to 0 the results have less variety and predefined properties achieved in more scenes. 
We use Jensen-Shannon divergence to quantify similarity between resulting distribution and uniform distribution. Uniform distribution is chosen as a perfect case for variety, the more the resulting distribution is close to the uniform distribution the more the results show diversity. In this paper we choose $\tau$ from the following condition:\\
\begin{align*}
\tau_{optimal} = \argmax_{0< \tau \leq 1} \min (V(\tau), W(\tau))\,,\\
V(\tau) = 1 - JSD(P^{\tau}, \mathcal{U})\,.
\end{align*}
Here, $V(\tau)$, $W(\tau)$ mean variety and percentage of successful scenes for fixed value of $\tau$ respectively. $JSD(P^{\tau}, \mathcal{U})$ is Jensen-Shannon divergence between resulting distribution and uniform distribution (the base 2 logarithm is used when calculating JSD, $0\leq JSD(P^{\tau}, \mathcal{U}) \leq 1$). Actions can be sampled by uniform distribution for several starting steps in the episode when it does not affect the accuracy. 
We use the PPO algorithm for several purposes: this algorithm uses both policy and value based updates, and it has been shown that this method is much faster than other reinforcement learning implementations like A2C \cite{mnih2016asynchronous}, A2C with trust region \cite{WangBHMMKF16} and TRPO \cite{SchulmanLMJA15}.  We also implemented our method with A3C \cite{mnih2016asynchronous}. In both cases the agent learned efficient policies, but PPO required much less iterations than A3C to converge. 

\section{Experiments and Results}
In order to evaluate our RLSS method, we address two problems: indoor planning as well as the generation of Angry Birds game levels. The initial setting is the same for all cases considered except for the number of operations.

\begin{figure}
\centering
\includegraphics[width=1.0\columnwidth]{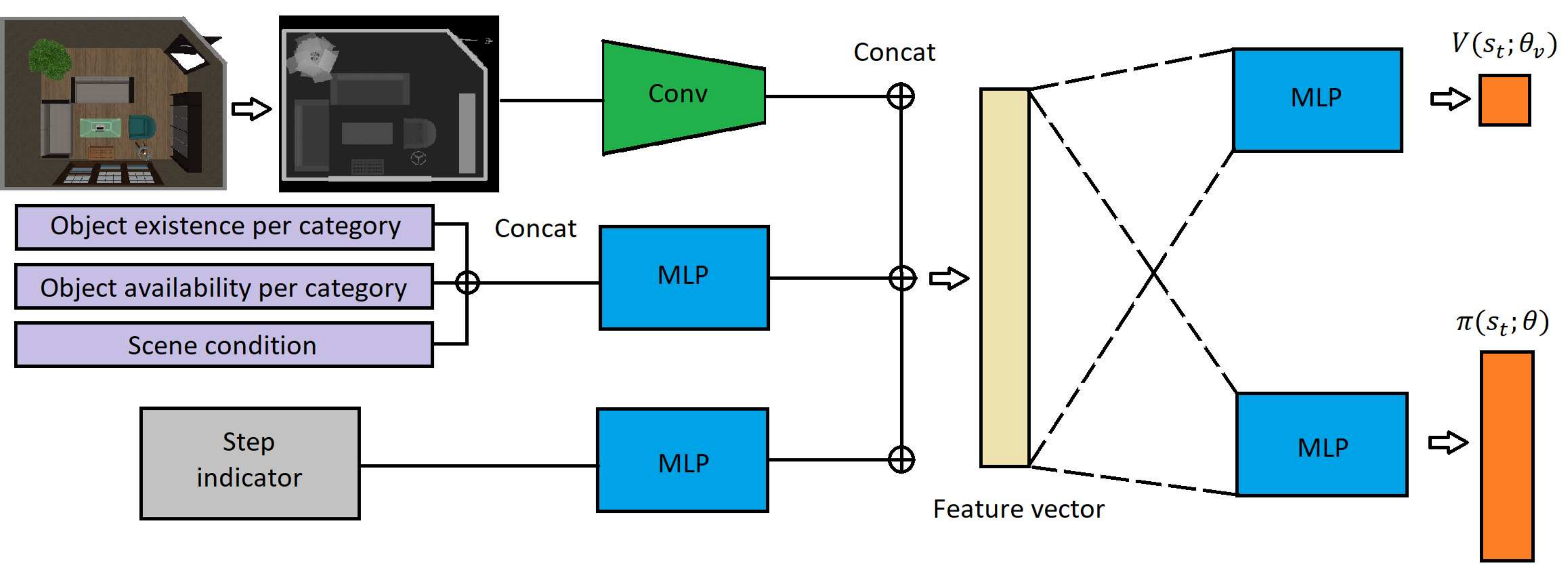}


\caption{Network architecture for our RLSS method. State includes top-down representation of the indoor scene, per category object existence and availability in the scene, scene condition and step indicator.}
\label{fig:nn_arch}
\end{figure}
\subsection{Implementation Details}

\textit{Framework.} We implement our method within the ChainerRL framework \cite{fujita2019chainerrl}. In order to use it, the problem we would like to solve must be formulated as an environment. We create our own environment within the Open AI Gym \cite{1606.01540} framework and add required functionalities to it.\\
\indent \textit{Network architecture and data representation.} The network architecture we use is illustrated in Fig.~\ref{fig:nn_arch}. MLP refers to fully connected layers, Conv specifies convolutional layers. Three convolutional layers are used for pre-processed input image and a fully connected layer for each of other two inputs, all of the layers are followed by ReLU nonlinearity \cite{DBLP:journals/corr/abs-1803-08375}. Then three outputs are concatenated to one feature vector. This is then continued by MLPs into two final outputs one for each of $V(s_t ; \theta_{v})$ value and $\pi(s_t ; \theta)$ policy. Each MLP consisted of two fully connected layers, first of which is followed by ReLU. The representation of the scene is consists of only the last frame. As a pre-processing, at first, we convert the scene to heightmaps \cite{article2, fisher2015activity}, a top-down depth rendered view of the scene, then downsize the image to $128 \times 128$ size. Object existence and availability are vectors consisting of $\{0, 1\}$ per object category, scene condition is normalized dividing by $R_{m}$ and is also represented as a vector. These three are concatenated and fed to MLP. Step indicator is represented as a one-hot encoding.\\
\indent \textit{Parameters.} We use a discount factor of $\gamma = 0.99$, the agents update of the network parameters after every $2048$ actions, minibatch size is equal to $64$ and number of epochs is equal to $10$. As for the optimization, we use Adam optimizer \cite{Adam} with a learning rate $lr = 3\cdot10^{-4}$. In order to stabilize the learning process we use a reward scaling with a factor of $rs = 10^{-2}$. In our experiments, the value loss coefficient is set to $c_1 = 1$ and the entropy regularization weight is equal to $c_2 = 0.01$.\\ 
\indent \textit{Training.} We have used non-parallel implementation with only a single actor-learner. The neural network training took around $13$ hours for $4$ million steps. In the last million steps we applied only greedy action sampling. 

\subsection{Indoor Planning}
\begin{figure}
\centering
\subfloat[Bedrooms]{%
  \includegraphics[width=0.8\columnwidth]{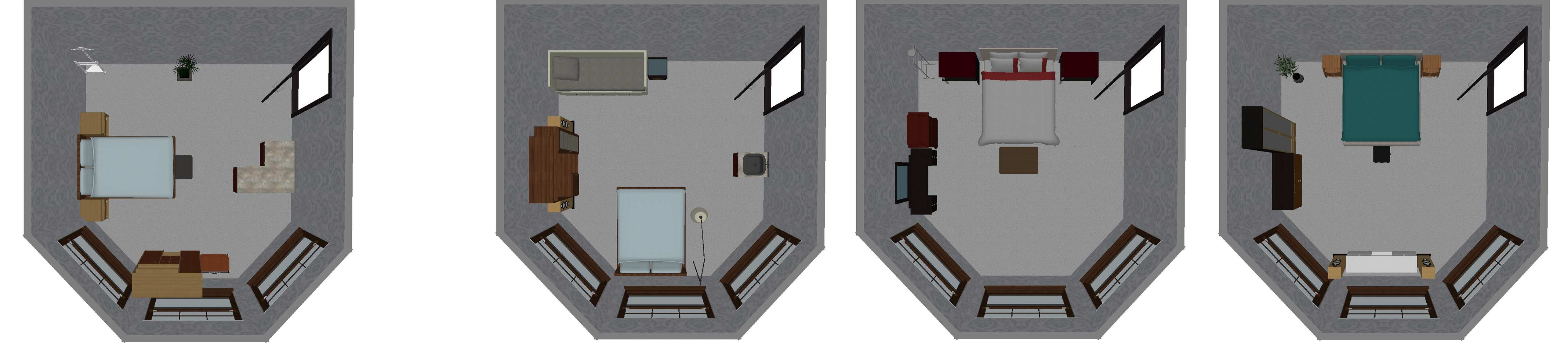}%
}

\subfloat[Living rooms]{%
  \includegraphics[width=0.8\columnwidth]{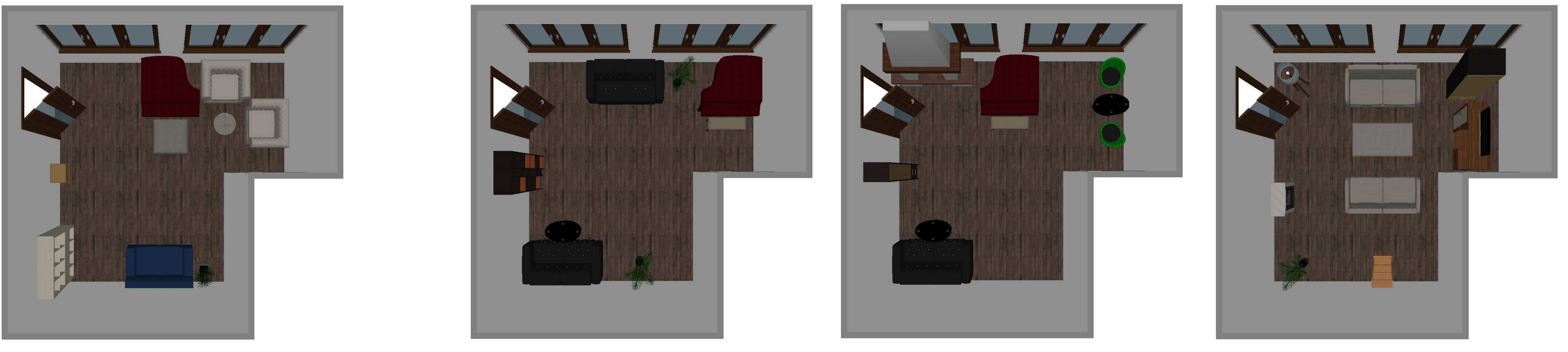}%
}

\subfloat[Offices]{%
  \includegraphics[width=0.8\columnwidth]{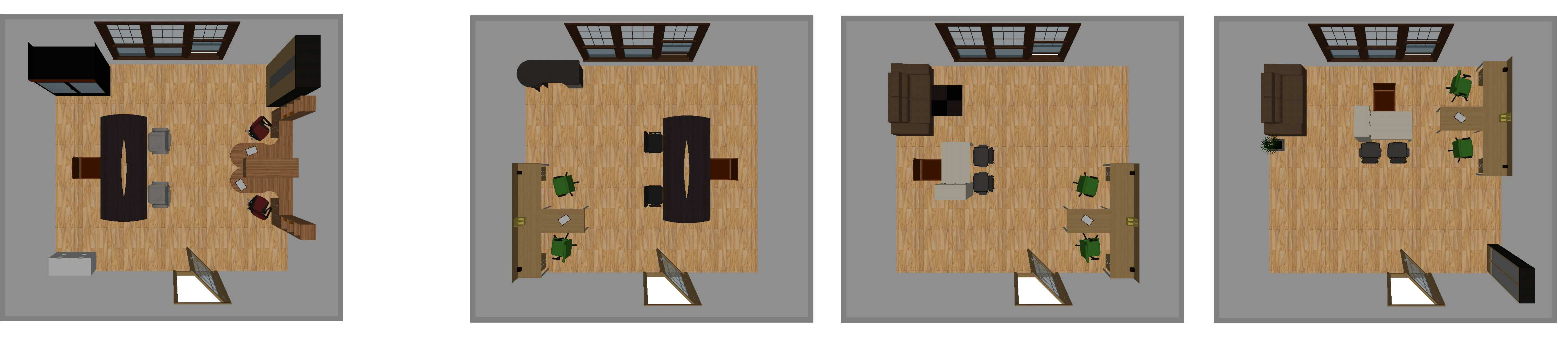}%
}
\caption{Illustration of the diversity of scenes generated with our RLSS method. The leftmost image in each column represents an example image, other images are the nearest neighbors amongst 1000 generated scenes. Scene similarity is measured using graph kernels \cite{articleFisher}.}
\label{fig:variety_comp}
\end{figure}

\begin{figure*}
\centering
\subfloat[Different bedroom scenes generated with Fast \& Flexible (left) and our RLSS method (right).]{%
  \includegraphics[clip,width=0.65\textwidth]{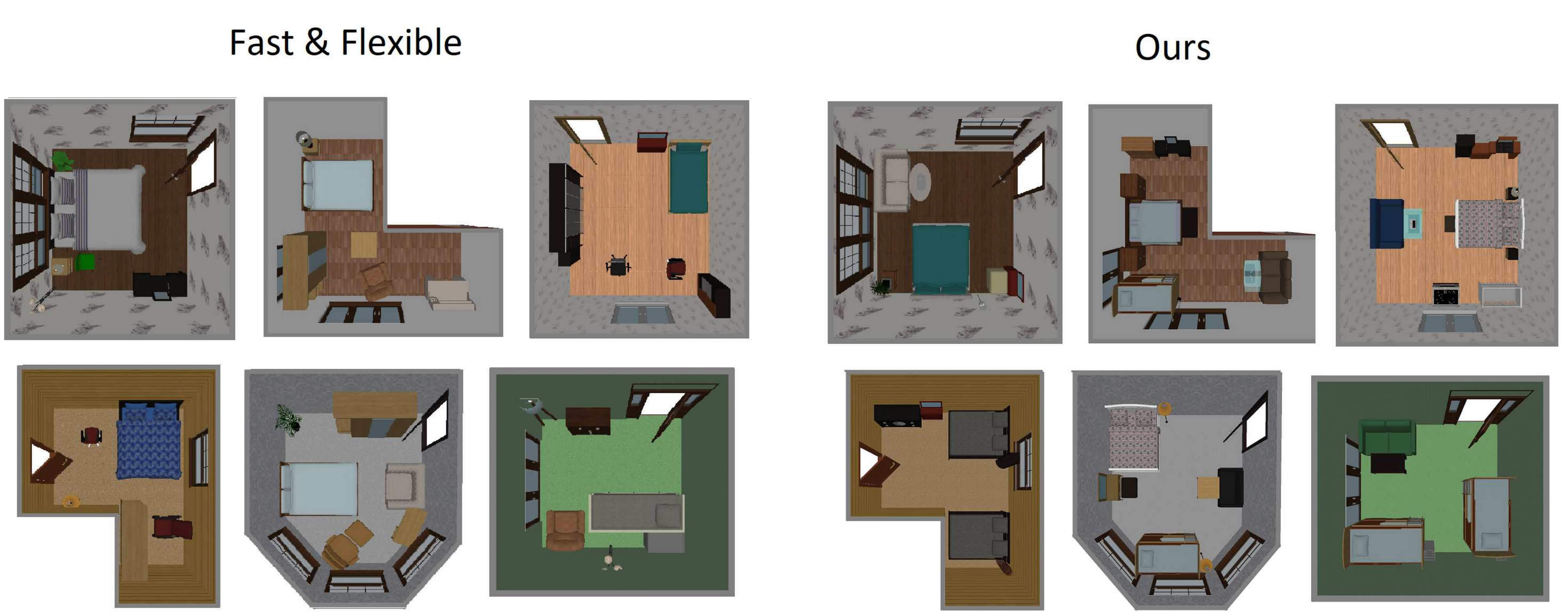}%
}

\subfloat[Different living room scenes generated with Fast \& Flexible (left) and our RLSS method (right).]{%
  \includegraphics[clip,width=0.65\textwidth]{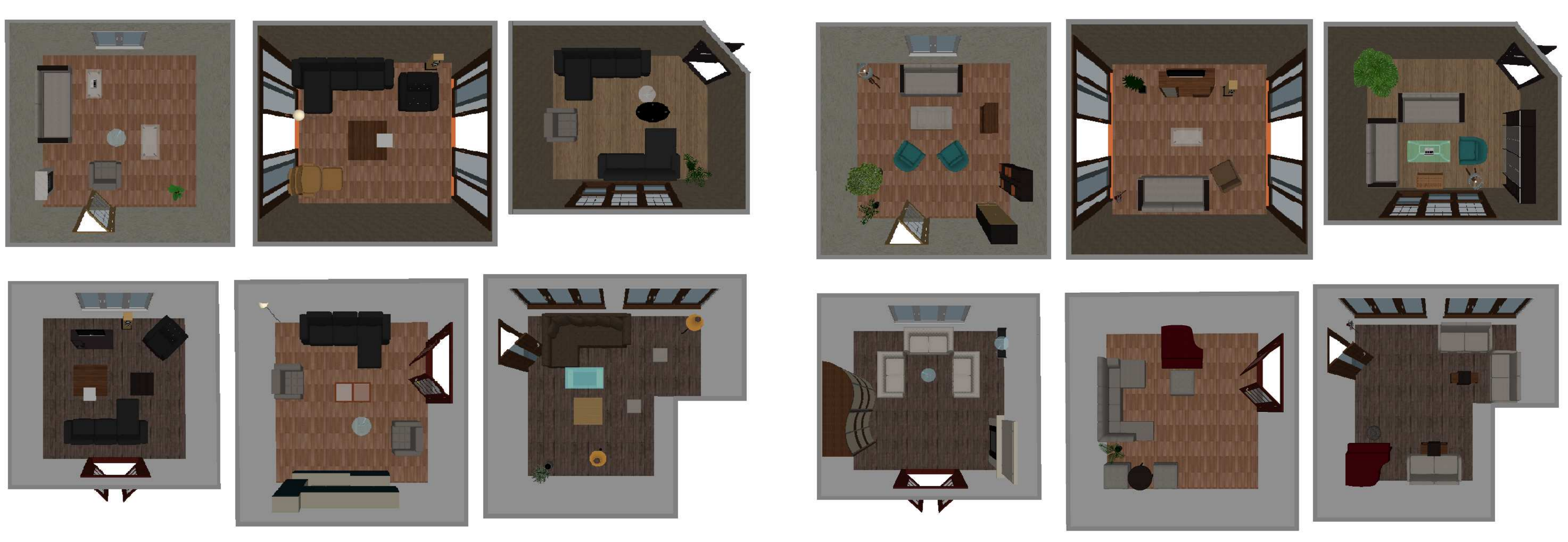}%
}

\subfloat[Different office scenes generated with Fast \& Flexible(left) and our RLSS method (right).]{%
  \includegraphics[clip,width=0.65\textwidth]{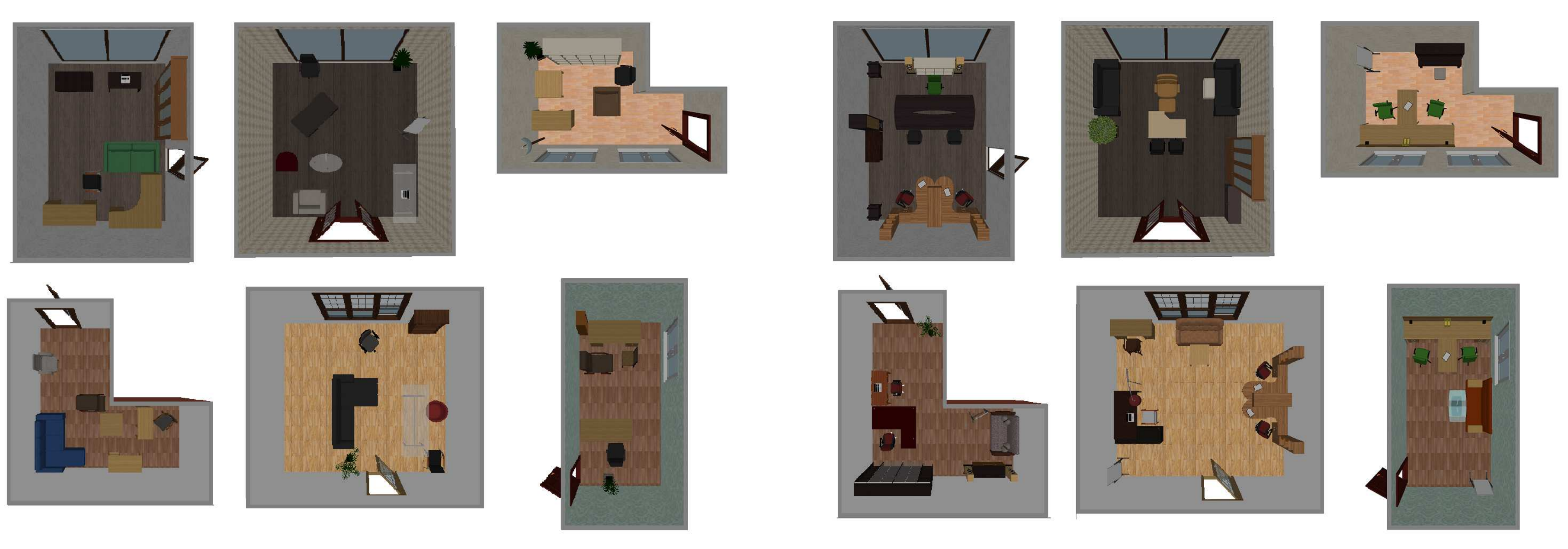}%
}
\caption{Qualitative comparison of scenes generated with Fast \& Flexible \cite{Ritchie_2019} and our method.}
\label{fig:method_comp}
\end{figure*}

In this problem, different room models are given, i.e.~the geometry of the room including walls, doors, floors, and ceilings. Three different types of rooms are considered: living rooms, bedrooms, and offices. Our task is to propose a generative method for placing objects, i.e.~furniture, within the room.\\
\indent Our approach is to group objects based on their functionalities, and place the current object to one of its appropriate group by analyzing existing not completed groups in the room. This helps us to generate rooms which are more similar to human planned ones. For example, some groups we use for synthesizing bedrooms are: (1) bed, nightstand, floor lamp, and ottoman; (2) desk and chair; (3) dressing table and ottoman. In order to find groups of objects that appear together in the room, we use human planned room models. We also analyze the arrangements, how the group objects come together with usual placing relative to a room architecture and instantiate objects according to this. Our representation to train the network combines a top-down view of the generated scene including positions of the door and the windows.\\
\indent We consider several hard constraints for this problem. First, two different objects should not share a common area (collision avoidance). Next, it should be possible for a human to move in the room conveniently. Finally, no object should block another object in the room, i.e.~all objects should be accessible for a human. We assign rewards for each complexity of structures and an additional reward for some important objects for this type of the room, e.g., placing a bed in the bedroom. The additional reward is assigned only once when the important object(s) is first time placed in the room during the current episode.\\
\indent \textit{Comparison.} To compare our work with the state-of-the-art, we consider recent convolutional neural network based approaches \cite{article2, article_grains, article_planit, Ritchie_2019} and an optimization based method \cite{articleMakeitHome}. We compare these methods on 60 randomly taken generated examples for each type of the room (bedrooms, living rooms, and offices) using several constraints including object-with-object and object-with-room boundary collisions, object accessibility and door blocking problems; see Table.~\ref{table:time_comp}. Our method handles these constraints much more accurately compared to other methods. However, given the wide variety and complexity of the room boundaries, our method also allows for a small percentage of constraint violations, especially with object accessibility related problems. For a fair comparison we applied our structure based rules on \cite{articleMakeitHome} and implemented their code as it was not available. It should be noted that \cite{articleMakeitHome} rearranges the selected set of objects in the room, which severely limits the variety. We also consider another baseline to evaluate the importance of learning. We apply Greedy Search to place objects in the scene with our rule based \textit{AssignReward()} function with and without rewards. In the first case objects are sampled uniformly, in the second case objects are sampled by rewards which are calculated for each object. Results show Table.~\ref{table:wolearning}, that our RLSS outperforms the baseline method by large margins. Scene complexity in this table specifies the maximal complexity of the structures present in the scene averaged over tests. In general, non-learning based baselines generate scenes for smaller rooms, but failed for large rooms, also reward calculation for each object results to an increase in synthesis time.\\
\begin{table*}[!t]
\centering
\begin{tabular}{|l|c|c|p{1.4cm}|p{1.4cm}|p{1.4cm}|c|}
\hline\noalign{\smallskip}
\multirow{2}{*}{{\bf Method}} & \multirow{2}{*}{Generation Time} & \multirow{2}{*}{Room Boundary} & \multicolumn{3}{|c|}{Acceptable scenes (by hard constraints)} & \multirow{2}{*}{KL divergence} \\ \cline{4-6} 
& & & Bedroom & Living & Office &\\
\noalign{\smallskip}
\hline
\noalign{\smallskip}
Deep Priors \cite{article2}  & $\sim$ 240\,sec. & {\bf any} & 60 \% & 83.3 \% & 61 \% & 0.89\\
GRAINS \cite{article_grains} & \textbf{0.1027\,sec.} & rectangular only & 64 \% & 80 \% & 52 \% & 0.83\\
Fast$\&$Flexible \cite{Ritchie_2019} & 1.858\,sec. & {\bf any} & 88.3 \% & 86.7 \% & 83.3 \% & 0.91\\
PlanIT \cite{article_planit} &  $\sim$ 72\,sec. & closed room boundaries & 80 \% & 83.3 \% & 80 \% & \textbf{0.72}\\
Make it Home \cite{articleMakeitHome} &  $\sim$ 22\,sec. & \textemdash  & 81 \% & 81 \% & \textemdash & \textemdash
\\
Ours& $\sim$ 1\,sec. & {\bf any} & \textbf{97.5} \% & \textbf{98} \% & \textbf{96} \% & 0.81
\\
\hline
\end{tabular}
\caption{Classification and performance comparison of different indoor scene generation methods. Better results or options are indicated with bold text in each column. \textemdash \, means no available results.}
\label{table:time_comp}
\end{table*}
\indent We find that handling different room boundaries is one of the fundamental limitations in indoor planning. Methods based on generating scenes for the rectangular room boundaries \cite{article_grains,Kermani2016,inproceedings_human_centric} or for rooms which should be encircled with walls \cite{article_planit} might work well for special case but might not generalize well for other cases. Moreover, using the same rendering for the synthesized results is important to compare results fairly. For these purposes, we compare our method with \cite{Ritchie_2019}. Visual comparison of the synthesized scenes with two different methods can be found in Fig.~\ref{fig:method_comp}.\\
\indent To evaluate the diversity of the generated scenes we consider graph kernels \cite{articleFisher}. Fig.~\ref{fig:variety_comp} shows similar scenes to the given example scene amongst 1000 generated scenes, for each type of the room. Scene similarity in this example is evaluated based on proximity and common function of objects (relationship) in the scene. Numerical assessment of the diversity of generated scenes based on Kullback Leibler (KL) divergence of object category distribution and uniform distribution is given in the Table~\ref{table:time_comp}.\\
\indent Moreover, we evaluated the performance of our RLSS method and related work as illustrated in Table~\ref{table:time_comp}. As illustrated there, our RLSS method is very competitive when with respect to its performance. Also, our approach is not data-driven, it does not require training data and is capable of handling different room boundaries.

\setlength{\tabcolsep}{1.4pt}

\begin{table}[]
\centering
\begin{tabular}{|l|p{0.8cm}|p{0.8cm}|p{0.8cm}|p{0.8cm}|p{1cm}|p{1.1cm}|}
\hline
\multirow{2}{*}{{\bf Method}} & \multicolumn{2}{l|}{Accuracy} & \multicolumn{2}{l|}{\begin{tabular}[c]{@{}l@{}}Structure\\  complexity\end{tabular}} & \multicolumn{2}{l|}{Synthesis time}\\
\cline{2-7}
& {\bf IP} & {\bf AB} & {\bf IP} & {\bf AB} & {\bf IP} & {\bf AB}\\
\hline\hline
GSearch w/o r & 10 \% & 4 \% & 1.5 & 2.7 & 8 sec. & 5 sec.\\
GSearch w r  & 25 \% & 19 \% & 2.3 & 2.8 & 67 sec. & 37 sec.\\
RLSS & \textbf{62} \% & \textbf{58} \% &\textbf{3.4} &\textbf{3.0} &\textbf{1 sec} &\textbf{0.5 sec.}\\ 
\hline
\end{tabular}
\caption{Comparison with Greedy Search with and without rewards. IP refers to Indoor Planning, AB specifies Angry Birds levels. Better results are indicated with bold.}
\label{table:wolearning}
\end{table}

\subsection{Level Generation}
\begin{figure}
\centering
\subfloat[MSG v2.0]{%
  \includegraphics[clip,width=0.8\columnwidth]{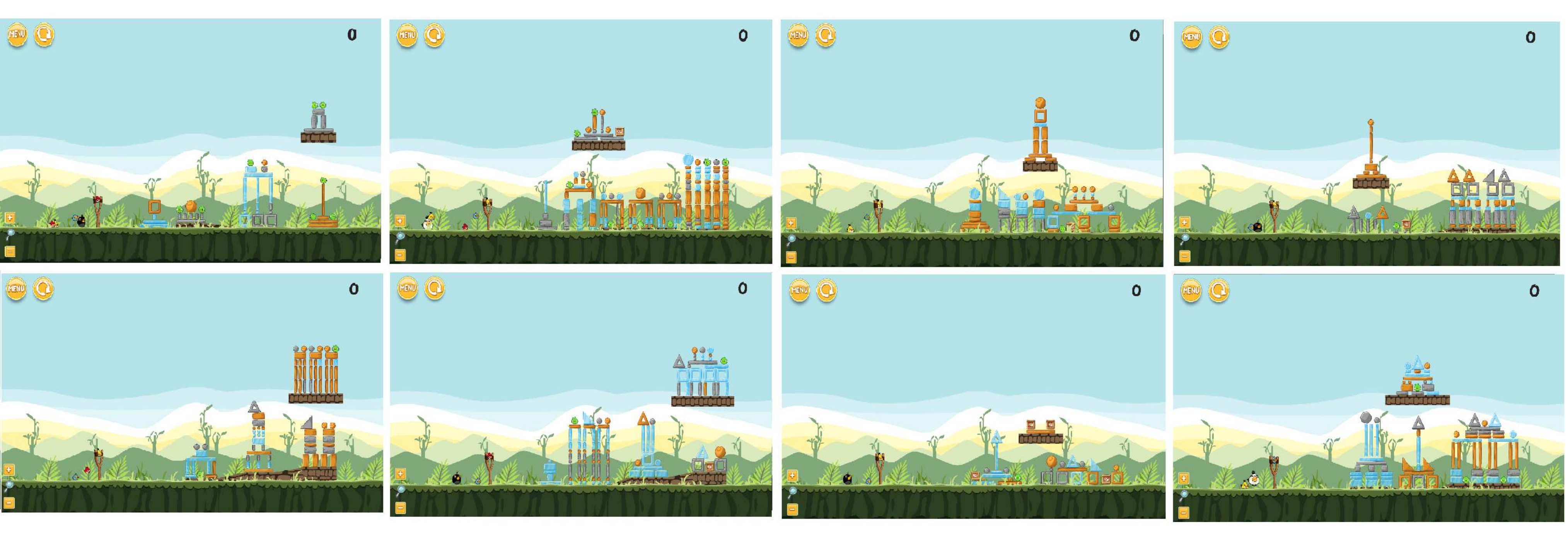}%
}

\subfloat[RLSS]{%
  \includegraphics[clip,width=0.8\columnwidth]{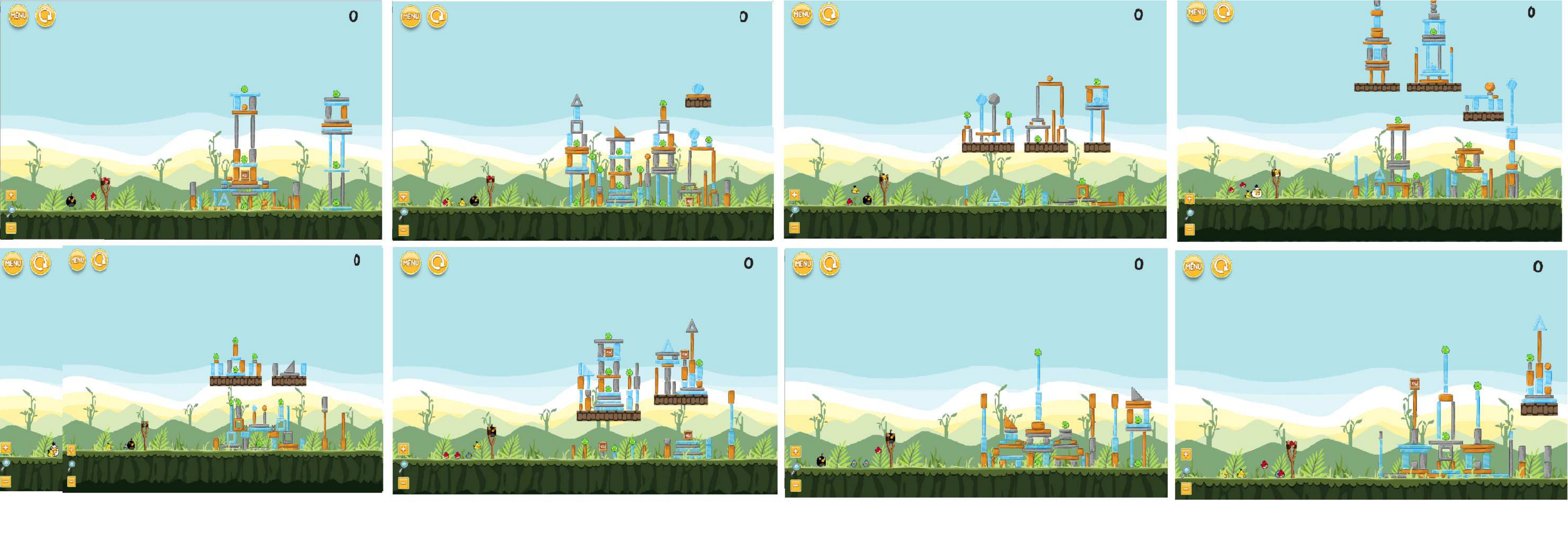}%
}

\caption{Illustration of several Angry Birds game levels generated with MSG v2.0 (top) and our RLSS method (bottom).}
\label{fig:angry_birds}
\end{figure}

As an additional benchmark from another context, we apply our RLSS method in order to generate levels for the physics-based game Angry Birds aiming for a wide variety of stable game levels. Blocks used in the game might differ by their size, shape, and material. Some blocks can be placed in the game environment in a horizontal as well as in a vertical position. All blocks can be classified as regular and irregular blocks. We build stable structures from regular blocks and add irregular blocks into these. All regular blocks have a box shape.\\
\indent The game area where we build stable blocks comprises a $800 \times 800$ pixel rectangle. Possible actions in this problem are to place a certain block in the scene. The last block is placed onto the top of the highest block in the given position. As a hard constraint we take stability.\\
\indent \textit{Comparison.} During testing, we maintain stable structures in the memory and build levels out of these. We compare our method with the MSG v2.0 winning generator for the 2017 and 2018 Angry Birds AI (level generation) competition \cite{article}. MSG v2.0 is based on generating the game content procedurally. This generator also generates levels from stable structures \cite{inproceedings}. We find that MSG v2.0 takes on average about 22 seconds to generate a single level, while our method performs this task in only about 0.5 seconds. A visual comparison of the generated levels is given in the Fig.~\ref{fig:angry_birds}. Levels rendered with Science Birds \cite{ferreira_2014_a}, an open-source, Unity-based clone of the Angry Birds game. KL divergence of object category distribution and uniform distribution for MSG v2.0 and our method is approximately equal to 0.23 and 0.21, respectively. The comparison results with the non-learning baselines introduced above are presented in the Table.~\ref{table:wolearning}.

\setlength{\tabcolsep}{1.4pt}

\begin{table}
\centering
\begin{tabular}{|l|c|c|c|}
\hline
{\bf Method} & Bedroom & Living & Office\\
\hline\hline
Deep Priors \cite{article2} & \textbf{76.2 ${\displaystyle \pm }$ 6.0} & \textbf{68.1 ${\displaystyle \pm }$ 10.0} & \textbf{81.4 ${\displaystyle \pm }$ 5.6} \\
GRAINS \cite{article_grains} & \textbf{72.3 ${\displaystyle \pm }$ 8.5} & \textbf{78.6 ${\displaystyle \pm }$ 9.1} & \textbf{80.1 ${\displaystyle \pm }$ 5.2} \\
Fast $\&$ Flexible \cite{Ritchie_2019} & 59.5 ${\displaystyle \pm }$ 10.8 & \textbf{67.6 ${\displaystyle \pm }$ 8.7} & \textbf{73.8 ${\displaystyle \pm }$ 8.5}\\
PlanIT \cite{article_planit} & \textbf{69.0 ${\displaystyle \pm }$ 9.5} &\textbf{68.1 ${\displaystyle \pm }$ 11.1} & \textbf{70.0 ${\displaystyle \pm }$ 7.5}\\ 
Make it Home \cite{articleMakeitHome} & \textbf{64.8 ${\displaystyle \pm }$ 9.4} &\textbf{61.4 ${\displaystyle \pm }$ 8.4} & \textemdash\\ 
\hline
\end{tabular}
\caption{Forced choice perceptual study results. Bold means our scenes are preferred with $95 \%$ confidence (${\displaystyle \pm }$ standard error), regular text means no preference. \textemdash \, means no available results. Higher is better.}
\label{table:forced_choice}
\end{table}
\subsection{Quantitative Evaluations}
To quantitatively evaluate results generated by our method we conducted a 2 alternative forced choice comparison on Amazon Mechanical Turk. Participants were asked to choose the most plausible scene out of two synthesized scenes placed side-by-side, one generated with our RLSS method and another one with other method. For indoor planning problem each participant performed overall 63 comparison task for each method, 20 for each room type and 1 for vigilance test. Images in these comparisons represent top-down view of a scene, rendered such that all objects are visible and colored with solid colors to help participants to choose scenes by the object arrangements not by colors or other not important factors. For each task we considered answers from 10 participants, who passed all the vigilance tests. The results of this perceptual study is summarized in the Table~\ref{table:forced_choice}. As it can be seen from the data, that our method shows the best results for almost all comparisons.\\
\indent We conducted similar perceptual study to compare generated Angry Birds levels, with 30 images and 15 participants. The result of this comparison show that our levels were preferred with \textbf{62.2 ${\displaystyle \pm }$ 6.9} (mean ${\displaystyle \pm }$ standard error) margins with $95 \%$ confidence.\\ \indent \textit{Ablation.} We train the neural network without Action Separation (ASE) \& Reward Designing (ASE+RD) and also without our Action Sampling (ASA) for Angry Birds level generation problem. The experiments show that Naive PPO does not learn to achieve the predefined design objectives, instead it learns to pile up box objects to maximize reward. PPO with ASE+RD and sampling the next action from the top 3 with $\epsilon$ exploration at the beginning ($\epsilon = 0.6$) does not demonstrate enough variety and accuracy Table~\ref{table:ablation}. 
\setlength{\tabcolsep}{1.4pt}

\begin{table}
\centering
\begin{tabular}{|l|c|c|}
\hline
{\bf Method} & Accuracy & KL divergence\\
\hline\hline
Naive PPO  & 0 \% & 1.5634 \\
PPO+ASE+RD & 28 \% & 0.6377 \\
RLSS & \textbf{58} \% & \textbf{0.2144}\\ 
\hline
\end{tabular}
\caption{Ablation study results: Accuracy and diversity of generated scenes, for our method and its variations. For the accuracy column, higher is better, while for the KL divergence column, lower is better.}
\label{table:ablation}
\end{table}


\section{Conclusion}

In this paper, we proposed a reinforcement learning algorithm for scene generation called RLSS. To the best of our knowledge, we are the first to propose such an algorithm that can produce a wide variety of scenes. To this end, we modify the state-of-the-art PPO reinforcement learning algorithm to sample from a large set of possible scenes.
We also propose a suitable solution for reward function design, which includes two components. Hard constraints describe scene attributes that have to be fulfilled and that are strictly enforced. Predefined design objectives describe design goals or scene configurations that make a scene more desirable.
Our results show that RLSS can produce a large variety of scenes with significantly higher quality than the current state of the art.

\subsection{Limitations and Future Work}
A limitation of our method is that it currently does not have a mechanism to combine learning from reward functions and scene examples at the same time. In our future work, if high quality scene data bases become available, we would like to investigate combinations of generative adversarial networks and reinforcement learning to tackle this challenging research problem. Also, extracting structures and implementing placement functions take additional time.\\
\indent In addition, we believe that a great research direction is to extend reinforcement learning to learn scene generation from given input images. One possible approach is to design reward functions that include an estimation of the similarity of generated scenes and images. While this approach is even more difficult than learning from scene examples, it has the advantage that image data sets are much easier to come by than scene data sets.
\section*{Acknowledgements}
This work was funded by KAUST through baseline funding. The valuable comments of the anonymous reviewers that improved the manuscript are gratefully acknowledged.

{\small
\bibliographystyle{ieee_fullname}
\bibliography{egbib}
}

\end{document}